\pdfoutput=1

\documentclass[11pt]{article}
\usepackage{emnlp2021}

\usepackage{tgtermes} 
\usepackage{latexsym}

\usepackage[T1]{fontenc}

\usepackage[utf8]{inputenc}

\usepackage{microtype,graphicx,booktabs,paralist}
\usepackage{amsmath,amsfonts,amssymb}

\newcommand*{\tran}{^{\mathsf{T}}}
\newcommand{\metric}[1]{\textit{#1}}
\newcommand{\rotmetric}[1]{\rotatebox{45}{\metric{#1}}}
\let\stress\emph
\let\sentence\textit
\newcommand\pair[1]{\textit{\textbf{#1}}}
\bgroup 
  \catcode`\_=\active
  \gdef\works{\catcode`_=\active
      \def_{\setbox0=\hbox{0}\hspace*{\wd0}\relax}}
\egroup

\title{Regressive Ensemble for Machine Translation Quality Evaluation}
\author{Michal~Štefánik \and Vít~Novotný \and Petr~Sojka \\
  Masaryk University, Faculty of Informatics, MIR group 
  \url{https://mir.fi.muni.cz}\\
  \texttt{\{stefanik.m,witiko\}@mail.muni.cz}, \texttt{sojka@fi.muni.cz}
}

\begin{document}
\hyphenation{Fast-Text Spear-man word-piece}
\maketitle

\begin{abstract}
This work introduces a simple regressive ensemble for evaluating machine translation quality based on a set of novel and established metrics. 
We evaluate the ensemble using a correlation to expert-based MQM scores of the WMT 2021 Metrics workshop.
In both monolingual and zero-shot cross-lingual settings, we show a significant performance improvement over single metrics.
In the cross-lingual settings, we also demonstrate that an ensemble approach is well-applicable to unseen languages.
Furthermore, we identify a strong reference-free baseline that consistently outperforms the commonly-used \metric{BLEU} and \metric{METEOR} measures and significantly improves our ensemble's performance.
\end{abstract}

\section{Introduction}

Automated evaluation of text generation is challenging due to many orthogonal qualitative aspects that the user expects from a text generation system.
In machine translation, we can observe errors of the so-called critical category \cite{wulczyn_thain_dixon_2017} such as hallucinating \cite{lee2019hallucinations}, omitting parts of the input from translation, or the negation of meaning \cite{matusov-2019-challenges}.

Consider an example of the last category:
\newline
\begin{compactitem}
    \item \sentence{Reference}: ``I never wrote this article, I just edited it.''
    \item \sentence{Hypothesis 1}: ``It is not my article, I just edited it.''
    \item \sentence{Hypothesis 2}: ``I never wrote this article, I never edited it.''
\end{compactitem}
\medskip

\noindent In this example, all \metric{BERTScore} \cite{zhang2019bertscore}, \metric{BLEUrt} \cite{sellam2020bleurt}, and \metric{Prism} \cite{thompson-post-2020-paraphrase} metrics rank \sentence{Hypothesis~2} higher than \sentence{Hypo\-thesis~1}.
In \metric{BLEUrt} and \metric{Prism}, this can be due to a known vulnerability of Transformers, which rely on a lexical intersection \cite{McCoy2019RightFT} if such a heuristic fits the problem sufficiently well.
Trivially, just counting the negations can easily remedy this specific problem.
However, such a heuristic would fail in many other cases, such as when we adjust \sentence{Hypothesis~1} to ``It is an article of somebody else, I just edited it.''.

Such cases motivate our ensemble approach that aims to expose both surface and deeper semantic properties of texts and subsequently learn to utilize these for the specific task of translation evaluation.
Even though the objective might constrain the particular metrics, data set, or systematically fail in some cases, another metric or their combination in the ensemble allows the flaw to be corrected.

\section{Metrics for machine translation evaluation}

This section reviews the related work, focusing on the metrics we used in our ensemble.

The standard and still widely-used surface-level metrics for the evaluation of machine translation quality are \metric{BLEU} \cite{papineni2002bleu}, \metric{ROUGE} \cite{lin-2004-rouge}, and \metric{TER} \cite{snover-etal-2006-study}.
Surface-level metrics are not able to capture the proximity of meaning in cases where one text paraphrases the other, which is an ability commonly observed in deep neural language models \cite{lewis2020bart}.
One metric that addresses this flaw is \metric{METEOR} \cite{banerjee-lavie-2005-meteor}, which utilizes WordNet to account for synonymy, word inflection, or token-level paraphrasing.

Evaluation of semantic text equivalence is closely related to a problem of accurate textual representations (embeddings).
The traditional method that we identify as relevant for the evaluation of segment-sized texts is FastText \cite{bojanowski2017enriching}.
FastText learns representations of character $n$-grams from which it creates a unified representation of tokens by averaging.
Additionally, a distance of a pair of texts can be computed directly from the token-level embeddings using methods such as the soft vector space model (the soft cosine measure, \metric{SCM}) \cite{novotny2018implementation}, or by solving a minimum-cost flow problem (the word mover's distance, \metric{WMD}) \cite{kusner2015from}.

Similar matching is performed by \metric{BERTScore} \cite{zhang2019bertscore}, which uses internal token embeddings of a selected BERT layer optimal for the task. 
Although the token representations are multilingual in some models \cite{devlin2018bert}, which makes \metric{BERTScore} usable without references, we are not aware of prior work evaluating it as such.
A possible drawback of cross-lingual alignment using a max-reference matching scheme of \metric{BERTScore} lies in a possibility of a significant mismatch of sub-word tokens in source and target text. 
In contrast, the metric that we refer to as \metric{WMD-contextual} uses the same embeddings as \metric{BERTScore} but uses the network-flow optimization matching scheme of \metric{WMD}.

Task-agnostic methods have recently been outperformed by methods that fine-tune a pre-trained model for a related objective: \metric{BLEUrt} \cite{sellam2020bleurt} fine-tunes a \metric{BERT} \cite{devlin2018bert} model directly on Direct Assessments of submissions to WMT to predict the judgements using a linear head over contextual embeddings of a classification \textit{[CLS]} token.
\metric{Comet} \cite{rei-etal-2020-comet} learns to predict Direct Assessments from tuples of source, reference, and translation texts with the triplet objective or the standard MSE objective.

Some of the most recent work incorporates latent objectives and/or data sets.
For instance, \metric{Prism} \cite{thompson-post-2020-paraphrase,thompson-post-2020-automatic} learns a language-agnostic representation from multilingual paraphrasing in 39~languages, thus being one of the few well-performing reference-free metrics.
The orthogonality of its training objective might lower its correlation to other methods that use contextual embeddings.

\section{Methodology}
\label{sec:methodology}

Our methodology aims to answer the following major question with additional supporting questions:

\begin{enumerate}
    \item[\textbf{1.}] \textbf{Can an ensemble of surface, syntactic, and semantic-level metrics significantly improve the performance of single metrics?}
    \item[2.] Can such an approach be applied cross-lingually, i.e., on languages that it has not been trained on? 
    \item[3.] Can surface-level metrics in reference-free configuration achieve results comparable to the reference-based ones?
    \item[4.] Are contextual token representations important for evaluating semantic equivalence, or can these be replaced with pre-inferred token representations?
\end{enumerate}

\subsection{Experimental setup}

We perform our primary evaluation on Multidimensional Quality Metrics (MQM) data set \cite{expert_errors}, where we use averaged judgement scores as our gold standard.
Where multiple judgements are available for the given pair of a source and a hypothesis, we average the scores over the judgements and consider this average as our gold standard.
We split the samples into train (80\%) and test (20\%) subsets based on unique source texts.

In our experimental framework, which we release as an open-source Python library and Docker image for ease of reproduction\footnote{\url{https://github.com/MIR-MU/regemt}}\footnote{\url{https://hub.docker.com/r/miratmu/regemt}}, we implement a selected set of the metrics based on their guidelines, together with a bunch of novel metrics, introduced in Section~\ref{sec:our_metrics} aiming to provide additional, orthogonal insight of textual equivalence. 

Subsequently, we train a regressive ensemble on the \stress{standardized} metric features of the whole train set, intending to predict the averaged MQM expert judgements.
We evaluate the ensemble, together with all other selected metrics using pairwise Spearman's rank correlation (Spearman's $\rho$) with the MQM judgements on the held-out 20\% test split.
\looseness=-1

In addition to our primary evaluation on the MQM data set, we perform our experiments on the Direct Assessments (DA) from WMT 2015 and 2016, and a dev set of Catastrophic errors from the Post-editing Dataset \cite{fomicheva2020mlqepe} of the Multilingual Quality Estimation Dataset (MLQE-PE) \cite{fomicheva-etal-2020-unsupervised} used for evaluation at the Quality Estimation workshop of WMT 2021. 
Refer to Section~\ref{sec:other_metrics} for a detailed description of our experiments on DA and MLQE-PE.

We performed all our evaluations on segment-level judgements.
To minimize the impact of calibration for each of the specific metrics to evaluation, we report Spearman's rank correlation coefficient, reflecting the mutual qualitative ordering rather than particular values of the judgements.

\subsection{Novel metrics}
\label{sec:our_metrics}

In addition to a selected set of metrics based on a literature review, we implement a set of novel metrics that allows our ensemble to reflect a wider variance of properties of the evaluated texts.

\subsubsection{Soft Cosine Measure}
\label{sec:scm}

The soft cosine measure (\metric{SCM}) \cite{novotny2018implementation} is the cosine similarity of texts in the soft vector space model, where the axes of terms are at an angle corresponding to their cosine similarity~$S$ in a token embedding space:
\begin{equation}
  \label{eq:scm}
  \text{SCM}(\vec{x}, \vec{y}) =
  \frac{
    \vec{x}\tran\cdot S\cdot \vec{y}
  }{
    \sqrt{\vec{x}\tran\cdot S\cdot \vec{x}} \cdot
    \sqrt{\vec{y}\tran\cdot S\cdot \vec{y}}
  }
\end{equation}
where $\vec{x}$ is a weighted bag-of-words (BoW) vector of a reference (or a source in a reference-free setting), and $\vec{y}$ is a weighted BoW vector of a hypothesis.
\looseness=-1

We use \metric{SCM} with two token representations:
\begin{enumerate}
\item We use the static token representations of FastText \cite{grave2018learning}.
We refer to the resulting metric as \metric{SCM}.
\item We use the contextual token representations of \metric{BERT} \cite{devlin2018bert} using the methodology of \metric{BERTScore} \cite{zhang2019bertscore}.
We collect representations of all tokens segmented by the WordPiece \cite{wordpiece-shorter} tokenizer, and we treat each unique (token, context) pair as a single term in our vocabulary.

Subsequently, we \stress{decontextualize} these representations as follows:
For each WordPiece token, we average the representations of all (token, context) pairs in the training corpus.
We refer to the resulting metric as \metric{SCM-decontextualized}.
Due to the multilingual character of the learned BERT token representations, this metric is applicable both in reference-based and source-based approaches.
\end{enumerate}

In addition to two token representations, we also use two different SMART weighting schemes of \citet{salton1988term} for the BoW vectors $\vec{x}, \vec{y}$ and the construction of the term similarity matrix~$S$:
\begin{enumerate}
\item We use raw term frequencies as \textit{weights} in the BoW vectors, the \texttt{nnx} SMART weighting scheme, and we construct the term similarity matrix in the vocabulary order.
We refer to the resulting metrics as \metric{SCM} and \metric{SCM-decontextualized}.
\item We use term frequencies discounted by inverse document frequencies as weights in the BoW vectors, the \texttt{nfx} SMART weighting scheme, and we construct the term similarity matrix in the decreasing order of inverse document frequencies \cite[Section 3]{novotny2018implementation}.
We refer to the resulting metrics as \metric{SCM-tfidf} and \metric{SCM-decontextualized-tfidf}.
\end{enumerate}

\subsubsection{Word Mover's Distance}
\label{sec:wmd}

The Word Mover's Distance (\metric{WMD}) \cite{kusner2015from} finds the minimum-cost flow~$F$ between vector space representations of two texts:
\begin{multline}
  \text{WMD}(\vec{x}, \vec{y}) = \text{minimum cumulative cost }
  F\tran\cdot S\\[0.6ex]
  \text{ subject to }
  \sum_j F_{ij}=x_i,
  \sum_i F_{ij}=y_j,
\end{multline}
where $\vec{x}$ is an $\ell_1$-normalized weighted BoW vector of a reference (or a source in reference-free setting), $\vec{y}$ is an $\ell_1$-normalized weighted BoW vector of a hypothesis, and $S$ is a term similarity matrix.

Similar to \metric{SCM} described in the previous section, we experiment with two token representations: FastText embeddings of whole tokens (\metric{WMD}) and decontextualized embeddings of WordPiece tokens (\metric{WMD-decontextualized}).
Additionally, we also use the contextual embeddings of WordPiece tokens (\metric{WMD-contextual}) to show the impact of decontextualization on the metric performance: If the impact is negligible, future work could avoid the costly on-the-fly inference of BERT representation and significantly reduce the vocabulary size.

Similarly to \metric{SCM}, we also use two different weighting schemes: raw term frequencies (\metric{WMD-*}) and term frequencies discounted by inverse document frequencies (\metric{WMD-*-tfidf}).

\subsubsection{Compositionality}

Our custom metric that we refer to as Compositionality constructs a transition graph of an arbitrary text $G_t$ based on directed, pairwise transitions of the tokens' part-of-speech (PoS) categories.
As the models for PoS tagging are language-dependent, we use the compliant---though not always systematically aligned---schemes of tagging used for training the taggers in English \cite{ontonotes}, German \cite{BrantsEtAl2002}, Chinese \cite{ontonotes}, and Norwegian \cite{unhammer2009rfr}.
\looseness=1

Subsequently, we row-normalise the values of matrix $G_t$ and we define a distance metric of \metric{Compositionality}~$C$ for $x = G_{t_1}$ and $y = G_{t_2}$:
\begin{equation}
C(x, y) = \ell_1\text{-norm}(x_{ii} - y_{ii}),
\end{equation}
where the PoS tags $i \in x$ and $i \in y$. 

In our submission, we apply this metric only if the language belongs to a set of the languages for which we have a tagger: English, German, Chinese, or Norwegian.
In reference-based evaluations, this constraint applies to the target language; in the case of reference-free evaluations, it applies to both source and target languages.

\subsection{Ensemble}
\label{sec:ensemble}

We ensemble the aforementioned metrics as predictors in a regression model, minimizing the residual between the average segment-level MQM scores and predicted targets.

We experiment with a wide range of simple regressors and observe a superior performance of simple approaches of fully connected, two-layer perceptron with 100-dimensional hidden layer and ReLu activation and linear regression with squared residuals.
We report the results for \metric{RegEMT} as the best-performing one of these classifiers picked on a 20\% held-out validation subset of the train data set. 

In addition to the ensemble of all available metrics, we evaluate a baseline regressive ensemble \metric{Reg-base} using solemnly two surface-level features: character-level source and target length according to WordPiece~\cite{wordpiece-shorter} tokens. 

\subsection{Cross-lingual experiments}

As expert judgments are incredibly costly to obtain, it is unrealistic to expect that the training data for the trained systems will be available in the future for a vast majority of language pairs containing under-resourced languages.
To estimate the performance of all metrics on uncovered language pairs, we perform a cross-lingual evaluation on average MQM judgements of two available language pairs: \pair{zh-en} and \pair{en-de}.

Where applicable, we fit the metric parameters on the train split of the non-reported language pair.
Subsequently, we evaluate and report the results on a test split of the reported pair to MQM judgements.

\subsection{Ablation study}
\label{sec:ablation}

To understand the impact of individual metrics in their roles as predictors for our ensemble, we use their pairwise correlations for systematic feature elimination.

In our ablation study, we iteratively select the metric with the highest Spearman's~$\rho$ to any other metric.
We eliminate the selected metric from our ensemble by fitting a new regression model on the remaining features.
We continue until all metrics are eliminated and evaluate the ensemble at each step of the process.

\subsection{Additional evaluations}
\label{sec:other_metrics}

To allow for additional insight into the consistency of the results to other relevant evaluation sources, together with an evaluation of the metrics in the novel application of critical error recognition, we perform the experiments analogically on a DA data set of the WMT submissions from years 2015 and 2016, as well as to the Critical Errors dev set of MLQE-PE data set for reference-free metrics.

In the case of DA judgements, we use the assessments from the year 2015 as a training split and assessments from the year 2016 as a test split.

In the case of MLQE-PE, we split the data analogically to MQM by splitting the unique source texts in an 80:20 ratio.
In this case, we consider as gold judgements the mean severity of error assigned by three annotators to each of the translations.

\section{Results}

\begin{table*}[t]
\centerline{\scalebox{0.85}{
\begin{tabular}{@{}l@{\hspace*{15pt}}l@{\hspace*{-10pt}}l@{\hspace*{0pt}}l@{\hspace*{-10pt}}l@{\hspace*{-17pt}}l@{\hspace*{-15pt}}l@{\hspace*{-23pt}}l@{\hspace*{-12pt}}l@{\hspace*{-20pt}}l@{\hspace*{-10pt}}l@{\hspace*{-12pt}}l@{\hspace*{-2pt}}l@{\hspace*{-2pt}}l@{\hspace*{-8pt}}l@{\hspace*{-2pt}}l@{\hspace*{-10pt}}l@{\hspace*{-8pt}}l@{\hspace*{-5pt}}l@{\hspace*{-10pt}}l@{\hspace*{-10pt}}l@{\hspace*{-10pt}}@{}}
\toprule
 & \rotmetric{RegEMT}
 & \rotmetric{Prism} 
 & \rotmetric{BERTScr} 
 & \rotmetric{WMD-cont} 
 & \rotmetric{WMD-dec} 
 & \rotmetric{WMD-dec-tf}
 & \rotmetric{SCM-dec} 
 & \rotmetric{SCM-dec-tf}
 & \rotmetric{Compos}
 & \rotmetric{Reg-base}
 & \rotmetric{Comet}
 & \rotmetric{SCM} 
 & \rotmetric{SCM-tf}
 & \rotmetric{WMD} 
 & \rotmetric{WMD-tf}
 & \rotmetric{BLEUrt}
 & \rotmetric{BLEU}
 & \rotmetric{METEOR} \\
\midrule
MQM-src \pair{zh-en}   &   \textbf{.59}     & .36   & .44     & .44       & .29      & .17          & .19      & .13          & .13    & .34       &       &     &         &     &         &        &      &        \\
MQM-src \pair{zh-en-X} & \textbf{.49}    & .36   & .44     & .44       & .29      & .17          & .19      & .13          & .13    & .34       &       &     &         &     &         &        &      &        \\
MQM-src \pair{en-de}   & \textbf{.36}    & .09   & .14     & .06       & .04      & .07          & .03      & .02          & .23    & .28       &       &     &         &     &         &        &      &        \\
MQM-src \pair{en-de-X} & \textbf{.31}    & .09   & .14     & .06       & .04      & .07          & .04      & .02          & .23    & .28       &       &     &         &     &         &        &      &        \\
\midrule
MQM-ref \pair{zh-en}   & \textbf{.62}    & .45   & .45     & .43       & .27      & .21          & .10      & .26          & .01    & .35       & .51   & .19 & .35     & .29 & .27     & .48    & .25  & .28    \\
MQM-ref \pair{zh-en-X} & \textbf{.62}    & .45   & .45     & .43       & .27      & .21          & .09      & .25          & .01    & .31       & .51   & .19 & .35     & .29 & .27     & .48    & .25  & .28    \\
MQM-ref \pair{en-de}   & \textbf{.60}    & .32   & .22     & .25       & .32      & .28          & .33      & .18          & .12    & .27       & .48   & .06 & .14     & .13 & .07     & .10    & .13  & .20    \\
MQM-ref \pair{en-de-X} & .38    & .32   & .22     & .25       & .32      & .28          & .34      & .17          & .12    & .29       & \textbf{.48}   & .06 & .14     & .13 & .07     & .10    & .13  & .20 \\
\midrule
DA 2016-src           & \textbf{.84}   & .72       & .74             & .73                   & .57                          & .51                   & .37                          & .51              & .29                  & .18   & .82 & .39        & .45 & .44        & .42    & .81  & .42    & .50 \\
DA 2016-tgt           & .68   & \textbf{.70}       & .34             & .25                   & .09                          & .10                   & .10                          & .24              & .13                  & .04   &     &            &     &            &        &      &        &     \\
catastrophic-src & \textbf{.29}   & .26       & .13             & .11                   & .12                          & .15                   & .13                          & .09              & .10                  & .09   &     &            &     &            &        &      &        &    \\
\bottomrule
\end{tabular}
}}
\caption{Results for Spearman's correlations with selected gold standards.
From top: correlation of source-based metrics (top) and reference-based metrics (middle) to averaged scores of MQM expert judgements for specified languages.
Results suffixed with \pair{X} are evaluated cross-lingually: both ensemble metrics are trained on other language pairs than evaluated.
(Bottom): Results for other data sets: Direct assessments of WMT 2016 submissions and dev set of Catastrophic translations from MLQE-PE data set; reported values are an average of correlations over all available language pairs.}
\label{tab:MQM_results}
\end{table*}

\paragraph{Correlations to MQM judgements.} Table~\ref{tab:MQM_results} lists correlations to MQM for source-based, i.e., reference-free metrics (upper) and reference-based metrics (middle).
Results reported for \metric{RegEMT} fit a selected regression model on the estimates of all the other metrics available for a given evaluation scheme.
As described in Section~\ref{sec:ensemble}, we pick the evaluated regression model based on its performance on a held-out portion of the train set: a two-layer perceptron for the source-based \pair{zh-en} pair and a simple linear regression in all other cases, with negligible mutual differences between regression models in performance on the validation set (below 2\%).
In the reports suffixed with \pair{X}, we fit the regression model on the other language pair than the one used for the evaluation.\looseness=-1

The results suggest that a simple regressive ensemble can benefit from the variance of the predictors in a majority of the evaluated configurations, including the cross-lingual settings and other evaluated datasets. 
We observe the highest margins in correlations in the case of MQM judgements.

\paragraph{Baseline ensemble.}
Table~\ref{tab:MQM_results} shows that \metric{Reg-base}, using only the counts of reference and hypothesis Word-pieces, demonstrates its consistent superiority over the standard surface-level metrics of \metric{BLEU} and \metric{METEOR}, even in the cross-lingual vs.\ monolingual comparison.
With respect to the MQM judgements, the correlations of \metric{Reg-base} are reasonably consistent; hence, in the reference-free cases of \pair{en-de} language pair, the correlation of \metric{Reg-base} is very close to the correlations of \metric{RegEMT}.

\paragraph{Importance of contextualization.}
The results in Table~\ref{tab:MQM_results} are inconsistent concerning the importance of contextualization in token-level metrics (\metric{WMD-cont*} vs.\ \metric{WMD-dec*} and \metric{SCM-cont*} vs.\ \metric{SCM-dec*}).
We observe a significant (15--16\%) decrease of correlation between a contextualized and decontextualized versions of WMD in \stress{all} cases of \pair{zh-en} language pair.
The situation differs in \pair{en-de} pair, where for the reference-based case, the correlation of decontextualized version of WMD is superior by~7\%.\looseness=-1

\paragraph{Metrics correlations.}
Table~\ref{tab:correlations} demonstrates mutual correlations of the evaluated metrics.
We see the strong pairwise correlations among the metrics based on contextualized representations, such as between \metric{Comet}, \metric{Prism}, \metric{BERTScore} and \metric{BLEUrt}; all of these are higher than $0.79$.
The situation is similar among the metrics based on static token representations of \metric{SCM} and \metric{WMD}, both with and without TF-IDF.

In contrast, we observe a low correlation of \metric{BLEU} and \metric{METEOR} to \metric{Reg-base} forming a cluster of surface-level metrics.

\begin{table*}[t]
\works\setlength{\tabcolsep}{3.3pt}
\centerline{\scalebox{0.88}{%
\begin{tabular}{@{}l@{\hspace*{5pt}}lllllllllllllllllll@{}}
\toprule
        & _0) & _1) & _2) & _3) & _4) & _5) & _6)
        & _7) & _8) & _9) & 10) & 11) & 12) & 13)
        & 14) & 15) & 16) & 17) & 18) \\
\midrule
_0) \metric{MQM avg scr}  & 1.   & .62  & .51 & .45  & .45  & .43  & .27  & .21  & .1   & .26  & .19  & .35  & .29  & .27  & .48  & .25  & .28  & .01    & .35  \\
\midrule
_1) \metric{RegEMT}  & .59  & 1.   & .56 & .5   & .44  & .42  & .25  & .23  & .4   & .28  & .15  & .36  & .3   & .3   & .55  & .26  & .32  & .11  & .51 \\
_2) \metric{Comet} &      &      & 1.  & .79  & .82  & .82  & .71  & .61  & .54  & .62  & .57  & .64  & .64  & .6   & .83  & .64  & .63  & .44 & .4 \\
_3) \metric{Prism}                        & .36  & .5   &     & 1.   & .9   & .9   & .85  & .77  & .63  & .71  & .71  & .77  & .79  & .77  & .8   & .81  & .77  & .54  & .19  \\
_4) \metric{BERTScr}                    & .44  & .48  &     & .13  & 1. & .13 & .06 & .01 & .11 & .1  & .13 & .04 & .76  & .73  & .82  & .77  & .74  & .56  & .18  \\
_5) \metric{WMD-cont}              & .44  & .5   &     & .53  & .88 & 1.   & .92  & .8   & .68  & .69  & .71  & .74  & .77  & .73  & .82  & .81  & .75  & .58  & .18  \\
_6) \metric{WMD-dec}        & .29  & .42  &     & .22  & .4  & .6  & 1.   & .89  & .81  & .7   & .77  & .72  & .77  & .77  & .75  & .81  & .71  & .66  & .37  \\
_7) \metric{WMD-dec-tf} & .17  & .3   &     & .08  & .12  & .32 & .69 & 1.   & .71  & .66  & .73  & .71  & .74  & .8   & .63  & .74  & .64  & .6   & .32  \\
_8) \metric{SCM-dec}        & .19  & .26  &     & .03  & .21  & .35 & .71 & .49 & 1.   & .51  & .67  & .49  & .55  & .58  & .57  & .61  & .53  & .65  & .58  \\
_9) \metric{SCM-dec-tf} & .13  & .26  &     & .2   & .32  & .43  & .47  & .26  & .29  & 1.   & .58  & .72  & .64  & .61  & .62  & .62  & .58  & .48  & .19  \\
10) \metric{SCM}                         &      &      &     &      &      &      &      &      &      &      & 1.   & .78  & .9   & .86  & .54  & .75  & .74  & .67  & .37  \\
11) \metric{SCM-tf}                  &      &      &     &      &      &      &      &      &      &      &      & 1.   & .85  & .84  & .6   & .71  & .72  & .52  & .13  \\
12) \metric{WMD}                          &      &      &     &      &      &      &      &      &      &      &      &      & 1.   & .94  & .6   & .81  & .83  & .6   & .17  \\
13) \metric{WMD-tf}                  &      &      &     &      &      &      &      &      &      &      &      &      &      & 1.   & .58  & .78  & .78  & .58  & .22  \\
14) \metric{BLEUrt}                      &      &      &     &      &      &      &      &      &      &      &      &      &      &      & 1.   & .6   & .58  & .43  & .11  \\
15) \metric{BLEU}                        &      &      &     &      &      &      &      &      &      &      &      &      &      &      &      & 1.   & .85  & .66  & .27  \\
16) \metric{METEOR}                      &      &      &     &      &      &      &      &      &      &      &      &      &      &      &      &      & 1.   & .56  & .15  \\
17) \metric{Compos}            & .13 & .11 &     & .13 & .06 & .01 & .11 & .1  & .13 & .04 &      &      &      &      &      &      &      & 1.   & .52  \\
18) \metric{Reg-base}        & .34 & .44 &     & .37 & .07 & .07 & .24 & .19 & .28 & .02 &      &      &      &      &      &      &      & .34 & 1.   \\
\bottomrule
\end{tabular}
}}
\caption{Pairwise Spearman's correlations of the evaluated metrics and their correlations to averaged MQM judgements for \pair{zh-en} language pair.
Top-right triangle: mutual correlations of reference-based metrics, bottom-left triangle: correlations of metrics supporting multilingual source-based evaluation.}
\label{tab:correlations}
\end{table*}

\begin{figure*}
	\centerline{\hbox to 1.04\textwidth{\hss\includegraphics[scale=.55]{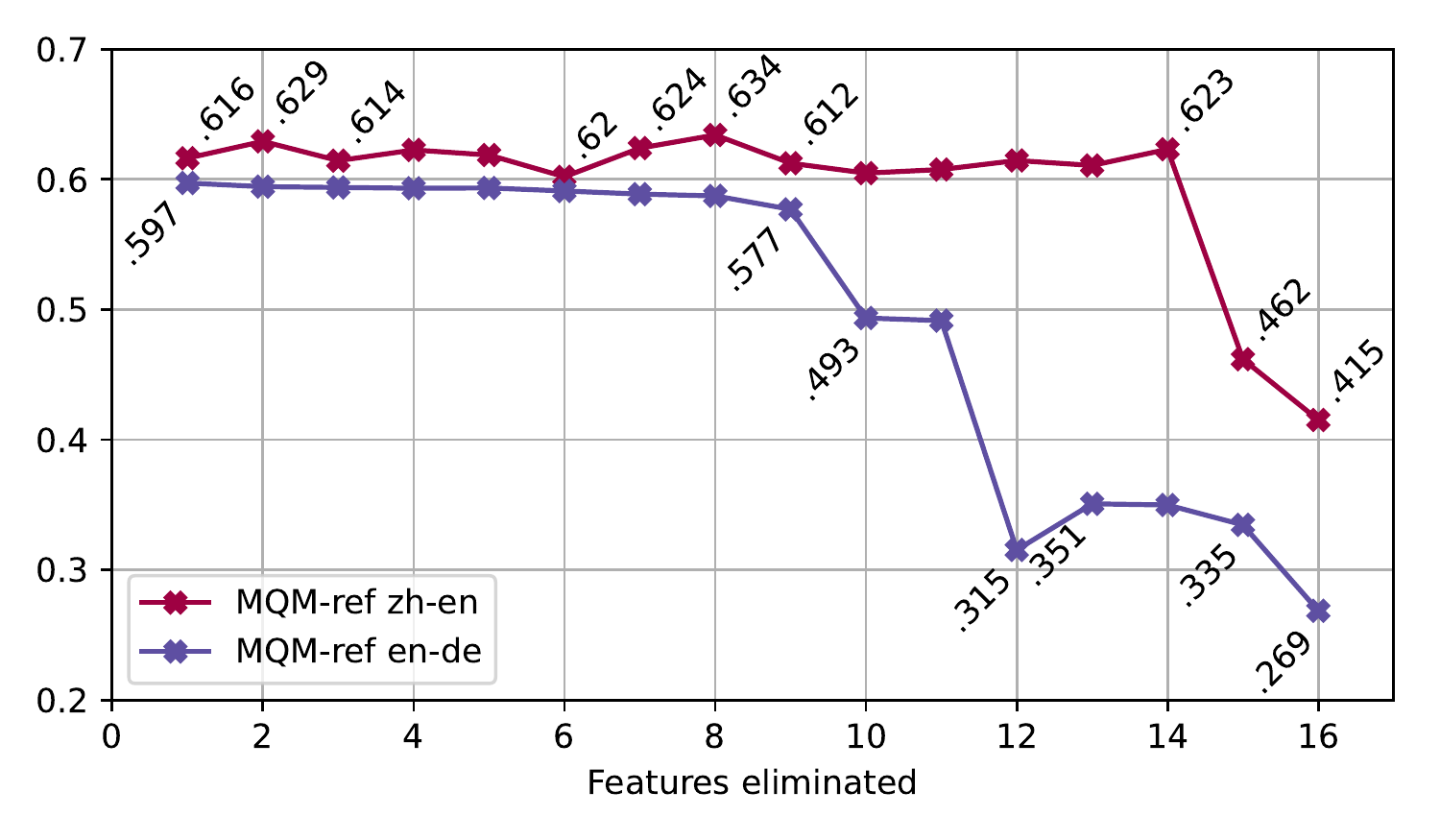}\hspace*{-2.7mm}%
	\includegraphics[scale=.55]{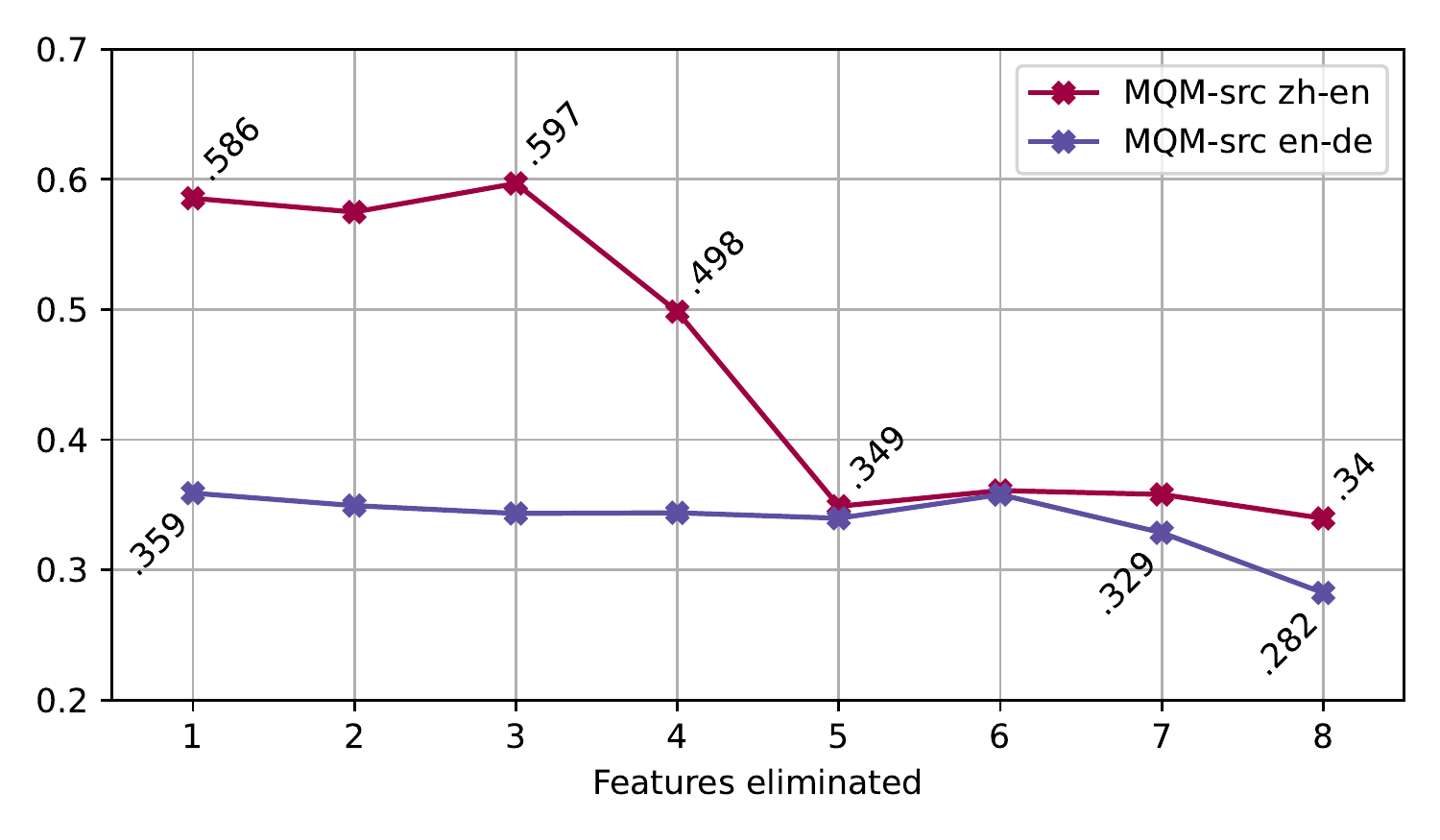}}~~~}
    \vspace*{-.6\baselineskip}
    \caption{Ablation study: correlation of ensemble estimator to averaged MQM judgements during incremental elimination of the most-correlated metric from the ensembled predictors.
    Left: ablation of reference-based metrics, right: ablation of source-based metrics.}
    \label{fig:correlations}
\end{figure*}

\paragraph{Ablation study.}
Figure~\ref{fig:correlations} displays performance development in Spearman's $\rho$ of the regressive ensemble when we incrementally eliminate the metrics from the set of ensembled predictors.
Following the methodology described in Section~\ref{sec:ablation}, the exact ordering of the metrics in ablation for \pair{zh-en} pair is shown in Table~\ref{tab:correlations}, and we observe it to be similar also for the other language pair of~MQM.

In ensembles of reference-based metrics (left), we observe a high consistency throughout the removal of most of the metrics.
A longer consistency in \pair{zh-en} case is attributed to a consistent performance in ensembling \metric{BLEUrt} (removed in step~14) and \metric{METEOR} (removed in step~15).
These metrics only reach the correlation of $0.48$ and $0.28$, respectively, when evaluated independently.
In \pair{en-de} case, the most significant drops can be attributed to a removal of the best-performing \metric{Comet} (step~9) and \metric{Prism} (step~11).

Ensemble of source-based metrics (right) shows significant drops in \pair{zh-en} pair after removing \metric{Prism} (step~3) and \metric{WMD-contextual} (step~4).
In \pair{en-de} language pair, the correlation is relatively low throughout the whole ablation process.
The least correlated and hence the last ones eliminated are \metric{WMD-decontextualized-tfidf} (step~6) and \metric{Prism} (step~7).

\section{Discussion}

\paragraph{Regressive ensemble.}
Following the objectives that we set in Section~\ref{sec:methodology}, we empirically confirm that an ensemble can push the quality of modeling the expert judgements in most of the configurations while performing close-to-the-best metrics on the others.
Additionally, we demonstrate that such ensemble is transferable to new language pairs and that its use is motivated by qualitative gains even in cross-lingual settings.

At the same time, one must acknowledge the limitations that an ensemble system exposes compared to single and unsupervised metrics.
An ensemble might inherit the systematic biases of each of its metrics.
This problem is observable in the results of the source-based \pair{en-de} pair of MQM in Table~\ref{tab:MQM_results}, where the ensemble follows the low correlations of its ensembled metrics.
Further, relying entirely on the metrics' consistency, the ensemble will inevitably expose errors in domains where some metrics behave markedly out of their usual range.

On the other hand, we argue that this might rarely be the case with the surface-level metrics that are mainly unsupervised.
We suspect it to be unlikely with learnable metrics, too, having their output space constrained by the range of their imitated metrics.

Values of correlations in Table~\ref{tab:correlations} and partially also the threshold metrics in Figure~\ref{fig:correlations} suggest that our ensemble relies primarily on trained contextualized metrics with regards to their correlation with the target as summarized in Table~\ref{tab:MQM_results}.
We suspect that oversampling of under-represented categories of errors would increase the significance of other types of metrics, as the under-represented error categories would be the ones where the fine-tuned metrics perform worse.

\paragraph{Baseline ensemble.}
Surprisingly, our baseline ensemble \metric{Reg-base} consistently outperforms other standard surface-based metrics such as \metric{BLEU}~\cite{papineni2002bleu}. This suggests possible applicability of surface-level metrics also in reference-free evaluation.

We suspect that the baseline features of length based on a multilingual WordPiece tokenizer~\cite{wordpiece-shorter} might reflect on the missing or inappropriately added segments more \textit{strictly} than other surface-level metrics.
At the same time, these errors are usually highly weighted in the overall score.

Table~\ref{tab:correlations} shows considerable orthogonality of \metric{Reg-base} to other metrics. This motivates the inclusion of the other weak surface metrics into the baseline ensemble to alleviate some of its apparent flaws.

\paragraph{Impact of contextualization.}
Based on the results of \metric{WMD-*} described in Section~\ref{sec:wmd}, one can not draw a consistent conclusion regarding the impact of contextualization.
On average, decontextualization has decreased the performance of \metric{WMD} by 8\%, but the original motivation of a significant improvement in the usability of estimators might compensate.
On the other hand, \metric{WMD-decontextualized} and \metric{WMD-decontextualized-tfidf} reached a considerable improvement of 16--18\% as compared to \metric{WMD} and \metric{WMD-tfidf} using FastText embeddings, while losing none of their flexibility.

\bigskip
\begin{raggedleft}
``It is the harmony of the diverse parts, their symmetry, their happy balance; in a word it is all that introduces order, all that gives unity, that permits us to see clearly and to comprehend at once both the ensemble and the details.\rlap{''}\\
\hfill Henri Poincar\'e
\end{raggedleft}

\vspace*{-.7\baselineskip}
\section{Conclusion}

This work evaluates the potential of ensembling multiple diverse metrics (\metric{RegEMT}) for an evaluation of machine translation quality and offers a new simple baseline metric \metric{Reg-base} that achieves better results than \metric{BLEU} and \metric{METEOR} by using just the source and reference lengths.
We measure significant gains in Spearman's correlation to MQM with \metric{RegEMT} compared to standalone metrics and we
demonstrate that even simple linear estimators can benefit from the expressivity that the methods of all levels of representation provide.
Additionally, as we demonstrate, the ensemble based on metrics supporting multilingualism can push the quality further even on unseen language pairs.

We recognize the inherent limitations of the regressive ensemble, which is inevitably slower, resource-heavier, and prone to inherit latent inductive biases of underlying metrics or their combinations.
However, \metric{RegEMT} shows the agility of the simple ensemble approach, which is in contrast to attempts to learn the full complexity of quality estimation through a single objective and allows the quality estimator to avoid the blind spots of particular metrics.
We hope that our results will motivate future work in the ensemble evaluation.

\bibliographystyle{acl_natbib}
\bibliography{stefanik}
\end{document}